\documentclass[journal]{IEEEtran}
\usepackage{graphicx}
\usepackage{epstopdf}
\usepackage{algorithm}
\usepackage{algorithmic}
\usepackage{color}
\usepackage{subfigure}
\usepackage{amsmath}

\usepackage{amssymb}
\usepackage{mathrsfs}

\begin{document}
\markboth{IEEE Transactions on Circuits and Systems for Video Technology}%
{}
%
\title{Cost-Effective Active Learning for Deep Image Classification}

%
%

\author{Keze~Wang, Dongyu~Zhang*, Ya~Li, Ruimao~Zhang, and Liang~Lin, {\em Senior Member, IEEE}
\IEEEcompsocitemizethanks{\IEEEcompsocthanksitem 
This work was supported in part by Guangdong Natural Science Foundation under Grant S2013050014548 and 2014A030313201, in part by State Key Development Program under Grant No 2016YFB1001000, and in part by the Fundamental Research Funds for the Central Universities. This work was also supported by Special Program for Applied Research on Super Computation of the NSFC-Guangdong Joint Fund (the second phase). We would like to thank Depeng Liang and Jin Xu for their preliminary contributions on this project. We gratefully acknowledge the support of NVIDIA Corporation with the donation of the Tesla K40 GPU used for this research.

Keze Wang, Dongyu Zhang, Ruimao Zhang and Liang Lin are with the School of Data and Computer Science, Sun Yat-sen University, Guang Zhou. E-mail: cszhangdy@163.com. (Corresponding author is Dongyu Zhang.)

Ya Li is with Guangzhou University, Guang Zhou. 

Copyright (c) 2016 IEEE. Personal use of this material is permitted.
However, permission to use this material for any other purposes must be obtained from the IEEE by sending an email to pubs-permissions@ieee.org.
}
}

\maketitle

\begin{abstract}
Recent successes in learning-based image classification, however, heavily rely on the large number of annotated training samples, which may require considerable human efforts. In this paper, we propose a novel active learning framework{, which is capable of building a competitive classifier with { optimal feature representation via} a limited amount of labeled training instances in an incremental learning manner. Our approach advances the existing active learning methods in two aspects. First, we incorporate deep convolutional neural networks into active learning. Through the properly designed framework, the feature representation and the classifier can be simultaneously updated with progressively annotated informative samples. Second, we present a { cost-effective} sample selection strategy to improve the classification performance with less manual annotations. Unlike traditional methods focusing on only the uncertain samples of low prediction confidence, we especially discover the large amount of high confidence samples from the unlabeled set { for feature learning.} 
Specifically, these high confidence samples are automatically selected and iteratively assigned pseudo-labels. We thus call our framework ``{ Cost-Effective Active Learning'' (CEAL)} standing for the two advantages.} Extensive experiments demonstrate that the proposed CEAL framework can achieve promising results on two challenging image classification datasets, i.e., face recognition on CACD database \cite{chen2014cross} and object categorization on Caltech-256 \cite{griffin2007caltech}.
\end{abstract}


\IEEEpeerreviewmaketitle


\section{Introduction}


{
Aiming at improving the existing models by incrementally selecting and annotating the most informative unlabeled samples, Active Learning (AL) has been well studied in the past few decades \cite{DBLP:journals/tgrs/DemirB15, DBLP:conf/icml/Brinker03, DBLP:journals/tkde/LongBCZIC15, Multi-classActiveLearning, DBLP:ALwithGaussian, DBLP:WhichFacestoTag, DBLP:journals/tit/CastroN08, DBLP:conf/cvpr/LiG13, DBLP:ConvexOptimization-AL, DBLP:A-Sequential-Algorithm}, and applied to various kind of vision tasks, such as image/video categorization \cite{DBLP:Multi-level-Adaptive-AL, DBLP:journals/tip/Zhang0C14, DBLP:conf/mmm/SunXJ15, DBLP:conf/icml/HoiJZL06, DBLP:journals/tgrs/TuiaRPKE10}, text/web classification \cite{DBLP:journals/jmlr/TongK01, DBLP:conf/icml/McCallumN98, DBLP:conf/icml/SchohnC00}, image/video retrieval \cite{DBLP:Large-scaleAL, DBLP:ExtremeVideoRetrieval}, etc. In the AL methods \cite{DBLP:journals/tgrs/DemirB15, DBLP:conf/icml/Brinker03, DBLP:journals/tkde/LongBCZIC15}, classifier/model is first initialized with a relatively small set of labeled training samples. Then it is continuously boosted by selecting and pushing some of the most informative samples to user for annotation. Although the existing AL approaches~\cite{CPAL13ICCV, DBLP:AdaptiveAL} have demonstrated impressive results on image classification, their classifiers/models are trained with hand-craft features (e.g., HoG, SIFT) on small scale visual datasets. The effectiveness of AL on more challenging image classification tasks has not been well studied. 

Recently, incredible progress on visual recognition tasks
has been made by deep learning approaches~\cite{krizhevsky2012imagenet, DBLP:conf/cvpr/CiresanMS12}. With sufficient labeled data~\cite{imagenet09CVPR}, deep 
convolutional neural networks(CNNs)~\cite{krizhevsky2012imagenet, vgg} are trained to directly learn features from raw pixels which has achieved the state-of-the-art performance for image classification. However in many real applications of large-scale image classification, the labeled data is not enough, since the tedious manual labeling process requires a lot of time and labor. Thus it has great practical significance to develop a framework by combining CNNs and active learning, which can jointly learn features and classifiers/models from unlabeled training data with minimal human annotations. 
But incorporating CNNs into active learning framework is not straightforward for real image classification tasks. This is due to the following two issues.

\begin{itemize}
\item The labeled training samples given by current AL approaches are insufficient for CNNs, as the majority unlabeled samples are usually ignored in active learning. AL usually selects only a few of the most informative samples (e.g., samples with quite low prediction confidence) in each learning step and frequently solicit user labeling. Thus it is difficult to obtain proper feature representations by fine-tuning CNNs with these minority informative samples. 

\item The process pipelines of AL and CNNs are inconsistent with each other. Most of AL methods pay close attention to model/classifier training. Their strategies to select the most informative samples are heavily dependent on the assumption that the feature representation is fixed. However, the feature learning and classifier training are jointly optimized in CNNs. Because of this inconsistency, simply fine-tuning CNNs in the traditional AL framework may face the divergence problem.
\end{itemize}
}

\begin{figure*}[htpb]
\begin{center}
\includegraphics[width=\textwidth]{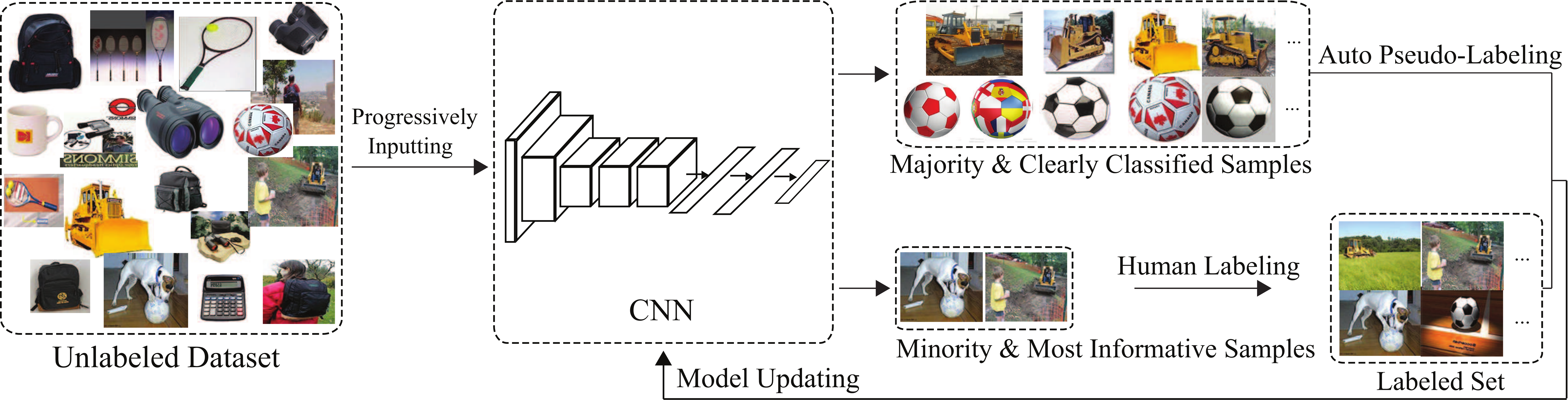}
\caption{Illustration of our proposed CEAL framework. Our proposed CEAL progressively feeds the samples from the unlabeled dataset into the CNN. Then both of the clearly classified samples and most informative samples selection criteria are applied on the classifier output of the CNN. After adding user annotated minority uncertain samples into the labeled set and pseudo-labeling majority certain samples, the model (feature representation and classifier of the CNN) are further updated.}
\label{fig:overview}
\end{center}
\end{figure*}

{
Inspired by the insights and lessons from a significant amount of previous works as well as the recently proposed technique, i.e., self-paced learning \cite{jiang2015self, zhao2015self, jiang2014easy, jiang2014self}, we address above mentioned issues by cost-effectively combining the CNN and AL via a complementary sample selection. In particular, we propose a novel active learning framework called ``Cost-Effective  Active Learning'' (CEAL), which is enabled to fine-tune the CNN with sufficient unlabeled training data and overcomes the inconsistency between the AL and CNN.

Different from the existing AL approaches that only consider the most informative and representative samples, our CEAL proposes to automatically select and pseudo-annotate unlabeled samples. As Fig.~\ref{fig:overview} illustrates, our proposed CEAL progressively feeds the samples from the unlabeled dataset into the CNN, and selects two kinds of samples for fine-tuning according to the output of CNN's classifiers. 
One kind is the minority samples with low prediction confidence, called most informative/uncertain samples. The predicted labels of samples are most uncertainty. For the selection of these uncertain samples, the proposed CEAL considers three common active learning methods: Least confidence \cite{Survey}, margin sampling \cite{DBLP:conf/ida/SchefferDW01} and entropy \cite{DBLP:journals/sigmobile/Shannon01}. The selected samples are added into the labeled set after active user labeling. The other kind is the majority samples with high prediction confidence, called high confidence samples. The predicted labels of samples are most certainty. For this certain kind of samples, the proposed CEAL automatically assigns pseudo-labels with no human labor cost.
As one can see that, these two kinds of samples are complementary to each other for representing different confidence levels of the current model on the unlabeled dataset. In the model updating stage, all the samples in the labeled set and currently pseudo-labeled high confidence samples are exploited to fine-tuning the CNN. 

The proposed CEAL advances in employing these two complementary kinds of samples to incrementally improve the model's classifier training and feature learning: the minority informative kind contributes to train more powerful classifiers, while the majority high confidence kind conduces to learn more discriminative feature representations.
On one hand, although the number is small, the most uncertainty unlabeled samples usually have great potential impact on the classifiers. Selecting and annotating them into training can lead to a better decision boundary of the classifiers. 
On the other hand, though unable to significantly improve the performance of classifiers, the high confidence unlabeled samples 
are close to the labeled samples in the CNN's feature space. Thus pseudo-labeling these majority high confidence samples for training is a reasonable data augmentation way for the CNN to learn robust features. 
In particular, the number of the high confidence samples is actually much larger than that of most uncertainty ones.  With the obtained robust feature representation, the inconsistency between the AL and CNN can be overcome. 

For the problem of keep the model stable in the training stage, many works \cite{kumar2010self, bengio2009curriculum} are proposed in recent years inspired by the learning process of humans that gradually include samples into training from easy to complex. Through this way, the training samples for the further iterations are gradually determined by the model itself based on what it has already learned \cite{jiang2014self}. In other words, the model can gradually select the high confidence samples as pseudo-labeled ones along with the training process. The advantages of these related studies motivate us to incrementally select unlabeled samples in a easy-to-hard manner to make pseudo-labeling process reliable. Specifically, considering the classification model is usually not reliable enough in the initial iterations, we employ high confidence threshold to define clearly classified samples and assign them pseudo-labels. When the performance of the classification model improves, the threshold correspondingly decreases.

}

The main contribution of this work { is threefold}. First, to the best of our knowledge, our work is the first one addressing the deep image classification problems in conjunction with active learning framework and convolutional neural networks training. Our framework can be easily extended to other similar visual recognition tasks. Second, this work also advances the active learning development, by introducing a {cost-effective} strategy to automatically select and annotate the high confidence samples, which improves the traditional samples selection strategies. Third, experiments on challenging CACD \cite{chen2014cross} and Caltech 256 \cite{griffin2007caltech} datasets show that our approach outperforms other methods not only in the classification accuracy but also in the reduction of human annotation.

The rest of the paper is organized as follows. Section II presents a brief review of related work. Section III discusses the component of our framework and the corresponding learning algorithm. Section IV presents the experiments results with deep empirical analysis. Section V concludes the paper.

\section{Related work}


The key idea of the active learning is that a learning algorithm should achieve greater accuracy with fewer labeled training samples, if it is allowed to choose the ones from which it learns \cite{Survey}. In this way, the instance selection scheme is becoming extremely important. One of the most common strategy is the uncertainty-based selection \cite{DBLP:A-Sequential-Algorithm, DBLP:journals/jmlr/TongK01}, which measures the uncertainties of novel unlabeled samples from the predictions of previous classifiers. In \cite{DBLP:A-Sequential-Algorithm}, Lewis \textit{et al.} proposed to extract the sample, which has the largest entropy on the conditional distribution over predicted labels, as the most uncertain instance. The SVM-based method \cite{DBLP:journals/jmlr/TongK01} determined the uncertain samples based on the relative distance between the candidate samples and the decision boundary. Some earlier works \cite{DBLP:CommitteeAlgorithm, DBLP:conf/icml/McCallumN98} also determined the sample uncertainty referring to a committee of classifiers (\textit{i.e. }examining the disagreement among class labels assigned by a set of classifiers). Such theoretically-motivated framework is called \textsl{query-by-committee }in literature \cite{Survey}. All of above mentioned uncertainty-based methods usually ignore the majority certain unlabeled samples, and thus are sensitive to outliers. The later methods have taken the information density measure into account and exploited the information of unlabeled data when selecting samples. These approaches usually sequentially select the informative samples relying on the probability estimation \cite{Multi-classActiveLearning, DBLP:ScalableAL-PAMI} or prior information \cite{DBLP:WhichFacestoTag} to minimize the generalization error of the trained classifier over the unlabeled data. For example, Joshi \textit{et al.} \cite{Multi-classActiveLearning} considered the uncertainty sampling method based on the probability estimation of class membership for all the instance in the selection pool, and such method can be effective to handle the multi-class case. In \cite{DBLP:WhichFacestoTag}, some context constraints are introduced as the priori to guide users to tag the face images more efficiently. At the same time, a series of works \cite{DBLP:AdaptiveAL, DBLP:ALwithGaussian} is proposed to take the samples to maximize the increase of mutual information between the candidate instance and the remaining unlabeled instances under the Gaussian Process framework. 
{ Li \textit{et al.} \cite{DBLP:AdaptiveAL} presented a novel adaptive active learning approach that combines an information density measure and a most uncertainty measure together to label critical instances for image classifications.} Moreover, the diversity of the selected instance over the certain category has been taken into consideration in \cite{DBLP:IncorporatingDiversity} as well. Such work is also the pioneer study expanding the SVM-based active learning from the \textit{single mode} to \textit{batch mode}. Recently, Elhamifar \cite{DBLP:ConvexOptimization-AL} \textit{et al.} further integrated the uncertainty and diversity measurement into a unified \textit{batch mode} framework via convex programming for unlabeled sample selection. Such approach is more feasible to conjunction with any type of classifiers, but not limited in max-margin based ones. It is obvious that all of the above mentioned active learning methods only consider those low-confidence samples (\textit{e.g.,} uncertain and diverse samples), but losing the sight of large majority of high confidence samples. We hold that due to the majority and consistency, these high confidence samples will also be beneficial to improve the accuracy and keep the stability of classifiers. Even more, we shall demonstrate that considering these high confidence samples can also reduce the user effort of annotation effectively.

\section{Cost-Effective Active Learning}
{
In this section, we develop an efficient algorithm for the proposed { cost-effective} active learning (CEAL) framework. Our objective is to apply our CEAL framework to deep image classification tasks by progressively selecting complementary samples for model updating. Suppose we have a dataset of $m$ categories and $n$ samples denoted as $D=\{x_i\}{_{i=1}^n}$. We denote the currently annotated samples of $D$ as $D^L$ while the unlabeled ones as $D^U$. The label of ${x}_i$ is denoted as ${y}_i=j, j\in\{1,...,m\}$, i.e., ${x}_i$ belongs to the $j$th category. We should give two necessary remarks on our problem settings. One is that in our investigated image classification problems, almost all data are unlabeled, i.e., most of \{$y_i$\} of $D$ is unknown and needed to be completed in the learning process. The other remark is that $D^U$ might possibly been inputted into the system in an incremental way. This means that data scale might be consistently growing. 

Thanks to {handling} both manually annotated and automatically pseudo-labeled samples together, our proposed CEAL model can progressively fit the consistently growing unlabeled data in such a holistic manner. The CEAL for deep image classification is formulated as follows:

\begin{equation}
\label{equ:goal}
\min_{\{\mathcal{W}, y_i, i \in D^U\}} -\frac{1}{n} \sum_{i=1}^n \sum_{j=1}^m \mathbf{1}\{y_i = j\}\log p( y_i =j \vert x_i; \mathcal{W}), 
\end{equation}
where $\mathbf{1}\{\cdot\}$ is the indicator function, so that $\mathbf{1}$\{a true statement\} = 1, and $\mathbf{1}$\{a false statement\} = 0, $\mathcal{W}$ denotes the network parameters of the CNN.  $p( y_i =j \vert x_i; \mathcal{W})$ denotes the softmax output of the CNN for the $j$th category, which represents the probability of the sample $x_i$ belonging to the $j$th classifiers.

The alternative search strategy is readily employed to optimize the above Eq.~(\ref{equ:goal}). Specifically, the algorithm is designed by alternatively updating the pseudo-labeled sample $y_i \in D^U$ and the network parameters $\mathcal{W}$. In the following, we introduce the details of the optimization steps, and give their physical interpretations. The practical implementation of the CEAL will also be discussed in the end.

\subsection{Initialization.} Before the experiment starts, the labeled samples $D^L$ is empty. For each class we randomly select few training samples from $D^U$ and manually annotate them as the starting point to initialize the CNN parameters $\mathcal{W}$.

\subsection{Complementary sample selection.} Fixing the CNN parameters $\mathcal{W}$, we first rank all unlabeled samples according to the common active learning criteria, then manually annotate those most uncertain samples and add them into $D^L$. For those most certain ones, we assign pseudo-labels and denote them as $D^H$. 

\textbf{Informative Sample Annotating:}
Our CEAL can use in conjunction with any type of common actively learning criteria, e.g., least confidence \cite{Survey}, margin sampling \cite{DBLP:conf/ida/SchefferDW01} and entropy \cite{DBLP:journals/sigmobile/Shannon01} to select $K$ most informative/uncertain samples left in $D^U$. The selection criteria are based on $p( y_i=j \vert x_i; \mathcal{W})$ which denotes the probability of $x_i$ belonging to the $j$th class. Specifically, the three selection criteria are defined as follows:
\begin{enumerate}
\item \emph{Least confidence:} 
Rank all the unlabeled samples in an ascending order according to the ${lc}_i$ value. ${lc}_i$ is defined as:
\begin{equation}
\label{eq:lc}
{lc}_i = \max_j p( y_i =j \vert x_i; \mathcal{W}).
\end{equation}
If the probability of the most probable class for a sample is low then the classifier is uncertain about the sample.
\item \emph{Margin sampling:} Rank all the unlabeled samples in an ascending order according to the ${ms}_i$ value. ${ms}_i$ is defined as:
\begin{equation}
\label{eq:ms}
{ms}_i = p( y_i =j_1 \vert x_i; \mathcal{W}) - p( y_i =j_2 \vert x_i; \mathcal{W}),
\end{equation}
where $j_1$ and $j_2$ represent the first and second most probable class labels predicted by the classifiers. The smaller of the margin means the classifier is more uncertain about the sample.
\item \emph{Entropy:} Rank all the unlabeled samples in an descending order according to their ${en}_i$ value. ${en}_i$ is defined as:
\begin{equation}\label{eq:entropy}
	{en}_i = -\sum_{j=1}^{m} p( y_i =j \vert x_i; \mathcal{W}) \log p( y_i =j \vert x_i; \mathcal{W}).
\end{equation}
This method takes all class label probabilities into consideration to measure the uncertainty. The higher entropy value, the more uncertain is the sample. 
\end{enumerate}

\textbf{High Confidence Sample Pseudo-labeling:} We select the high confidence samples from $D^U$, whose entropy is smaller than the threshold $\delta$. Then we assign clearly predicted pseudo-labels to them. The pseudo-label $y_i$ is defined as:
\begin{equation}\label{equ:pseudo}
\begin{gathered}
j^* = \arg \max_{j} p( y_i=j \vert x_i; \mathcal{W}), \\
y_i = \left\{
\begin{array}{c}
 j^*, \\
0,%
\end{array}
\begin{array}{c}
en_i < \delta, \\
\text{otherwise.}
\end{array}
\right.
\end{gathered}	
\end{equation}
where $y_i$ = 1 denotes that $x_i$ is regarded as high confidence sample. The selected samples are denoted as $D^H$. Note that compared with classification probability  $p( y_i=j^* \vert x_i; \mathcal{W})$ for the $j^*$th category, the employed entropy $en_i$ holistically considers the classification probability of the other categories, i.e., the selected sample should be clearly classified with high confidence. The threshold $\delta$ is set to a large value to guarantee a high reliability of assigning a pseudo-label.

\begin{algorithm}[htb]
\caption{ Learning Algorithm of CEAL}
\label{alg:DALFramwork}
\begin{algorithmic}[1]
\REQUIRE ~~\\
Unlabeled samples ${D}^U$, initially labeled samples ${D}^L$, uncertain samples selection size $K$, high confidence samples selection threshold $\delta$, threshold decay rate $dr$, maximum iteration number $T$, fine-tuning interval $t$.
\ENSURE ~~\\
CNN parameters $\mathcal{W}$.

\STATE { Initialize $\mathcal{W}$ with ${D}^L$.}
\WHILE{ { not reach maximum iteration $T$}}
\STATE {Add $K$ uncertainty samples into $D^L$ based on Eq.~(\ref{eq:lc}) or (\ref{eq:ms}) or (\ref{eq:entropy}), }
\STATE {Obtain high confidence samples $D^H$ based on  Eq.~(\ref{equ:pseudo})}
\STATE {In every $t$ iterations:}\begin{itemize}\setlength{\itemindent}{1em}
\item {Update $\mathcal{W}$ via fine-tuning according to Eq.~(\ref{equ:finetune}) with $D^H \cup D^L$}
\item {Update $\delta$ according to Eq.~(\ref{equ:thresh})}
\end{itemize}
\ENDWHILE
\RETURN $\mathcal{W}$
\end{algorithmic}
\end{algorithm}

\subsection{CNN fine-tuning} Fixing the labels of self-labeled high confidence samples $D^H$ and manually annotated ones $D^L$ by active user, the Eq.~(\ref{equ:goal}) can be simplified as:
\begin{equation}
\label{equ:finetune}
\begin{gathered}
\min_{\mathcal{W}} -\frac{1}{N}\sum_{i = 1}^N \sum_{j=1}^m \mathbf{1}\{y_i = j\}\log p( y_i =j \vert x_i; \mathcal{W}), \\
\end{gathered}
\end{equation}
where $N$ denotes the number of samples in $D^H \cup D^L$.
We employ the standard back propagation to update the CNN's parameters $\mathcal{W}$. Specifically, let $\mathcal{L}$ denote the loss function of Eq.~(\ref{equ:finetune}), then the partial derive of the network parameter $\mathcal{W}$ according to Eq.~(\ref{equ:finetune}) is:
\begin{equation}\label{eq:partial}
\begin{split}
\frac{\partial \mathcal{L}}{\partial \mathcal{W}} &= \partial \frac{ -\frac{1}{N} \sum_{i=1}^{N} \sum_{j=1}^m \mathbf{1}\{y_i = j\}\log p( y_i =j \vert x_i; \mathcal{W})}{ \partial \mathcal{W}}  \\
&=  -\frac{1}{N} \sum_{i=1}^{N} \sum_{j=1}^m \mathbf{1}\{y_i = j\} \partial \frac{ \log p( y_i =j \vert x_i; \mathcal{W})}{ \partial \mathcal{W}}  \\
&= - \frac{1}{N}\sum_{i=1}^{N} (\mathbf{1}\{y_i=j\} - p(y_i=j|x_i;\mathcal{W}))\frac{\partial z_j(x_i;\mathcal{W})}{\partial \mathcal{W}},
\end{split}
\end{equation}
where $\{z_j(x_i;\mathcal{W})\}_{j=1}^m$ denotes the activation for the $i$th sample of the last layer of CNN model before feeding into the softmax classifier, which is defined as:
\begin{equation}
p( y_i =j \vert x_i; \mathcal{W}) = \frac{e^{z_j(x_i; \mathcal{W} )}}{\sum_{t=1}^m e^{z_t(x_i; \mathcal{W})}}
\end{equation}

After fine-tuning we put the high confidence samples $D^H$ back to ${D}^U$ and erase their pseudo-label.

\subsection{Threshold updating}
As the incremental learning process goes on, the classification capability of classifier improves and more high confidence samples are selected, which may result in the decrease of incorrect auto annotation. In order to guarantee the reliability of high confidence sample selection, at the end of each iteration $t$, we update the high confidence sample selection threshold by setting 
\begin{equation}\label{equ:thresh}
\delta = \left\{
\begin{array}{c}
 \delta_0, \\
\delta - dr * t,%
\end{array}
\begin{array}{c}
t = 0, \\
t > 0.
\end{array}
\right.
\end{equation}
where $\delta_0$ is the initial threshold, $dr$ controls the threshold decay rate. 

\begin{figure*}[!ht]
\begin{center}
\includegraphics[width=\textwidth]{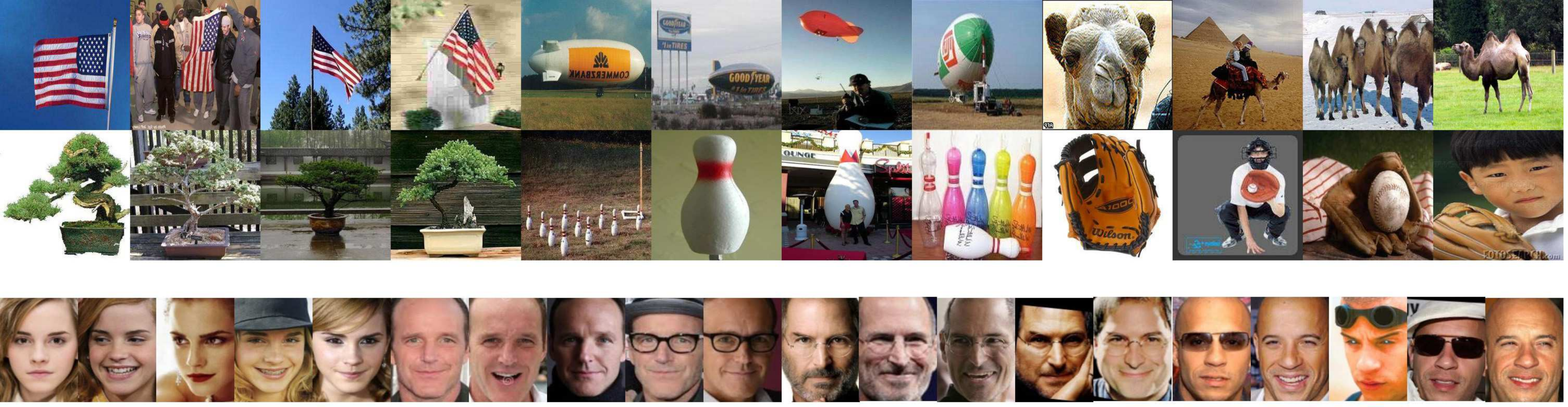}
\vspace{-8mm}
\caption{We demonstrate the effectiveness of our proposed heuristic deep active learning framework on face recognition and object categorization. The first and second line: sample images from the Caltech-256 \cite{griffin2007caltech} dataset. The last line: samples images from the Cross-Age Celebrity Dataset \cite{chen2014cross}. }
\label{fig:xintu}
\end{center}
\end{figure*}

The entire algorithm can be then summarized into Algorithm ~\ref{alg:DALFramwork}. It is easy to see that this alternative optimizing strategy finely accords with the pipeline of the proposed CEAL framework.

%
%
}

%
%
%
%
%

\section{Experiments}

\subsection{Datasets and Experiment settings}
\subsubsection{Dataset Description}
In this section, we evaluate our cost-effective active learning framework on two public challenging benchmarks, i.e., the Cross-Age Celebrity face recognition Dataset (CACD) \cite{chen2014cross} and the Caltech-256 object categorization \cite{griffin2007caltech} Dataset (see Figure~\ref{fig:xintu}). CACD is a large-scale and challenging dataset for face identification and retrieval problems. It contains more than $160,000$ images of $2,000$ celebrities, which are varying in age, pose, illumination, and occlusion.
Since not all of the images are annotated, we adopt a subset of $580$ individuals from the whole dataset in our experiments, in which, $200$ individuals are originally annotated and $380$ persons are extra annotated by us.
Especially, $6,336$ images of $80$ individuals are utilized for pre-training the network and the remaining $500$ persons are used to perform the experiments. Caltech-256 is a challenging object categories dataset. It contains a total of $30,607$ images of $256$ categories collected from the Internet.

\subsubsection{Experiment setting}
For CACD, we utilize the method proposed in \cite{xiong2013supervised} to detect the facial points and align the faces based on the eye locations. We resize all the faces into $200 \times 150$, then we set the parameters: $\delta_0 = 0.05$, $dr = 0.0033$ and $K = 2000$. For Caltech-256, we resize all the images to $256 \times 256$ and we set $\delta_0 = 0.005$, $dr = 0.00033$ and $K = 1000$. Following the settings in the existing active learning method \cite{DBLP:ConvexOptimization-AL}, we randomly select $80\%$ images of each class to form the unlabeled training set, and the rest are as the testing set in our experiments. Among the unlabeled training set, we randomly select $10\%$ samples of each class to initialize the network and the rest are for incremental learning process. To get rid of the influence of randomness, we average 5 times execution results as the final result.

\begin{table}[ht]
\begin{center}
\caption{ The detailed configuration of the CNN architecture used in CACD \cite{chen2014cross}. It takes the $200\times150\times3$ images as input and generates the 500-way softmax output for classes prediction. The ReLU \cite{nair2010rectified} activation function is not shown for brevity.}
\label{table:cacd_cnn}
\begin{tabular}{|c|c|c|} \hline
		layer type 			&   kernel size/stride 		&    output size	\\
	\hline
		convolution 			& $5\times5/2$ 				&  $98\times73\times32$  \\
		max pool    			& $3\times3/2$ 				&  $48\times36\times32$  \\	
		LRN 					&              				&  $48\times36\times32$  \\
		convolution(padding2)		& $5\times5/1$ 				&  $48\times36\times64$  \\
		max pool 				& $3\times3/2$ 				&  $23\times17\times64$  \\
		LRN					& 						&  $23\times17\times64$  \\
		convolution(padding1)		& $3\times3/1$ 				&  $23\times17\times96$  \\
		fc(dropout$50\%$) 		&  						&  $1\times1\times1536$  \\
		fc(dropout$50\%$) 		&  						&  $1\times1\times1536$  \\		
		softmax 				&              				&  $1\times1\times500$  \\
	\hline
\end{tabular}
\end{center}
\end{table}

\begin{table}[ht]
\begin{center}
\caption{ The detailed configuration of the CNN architecture used in Caltech-256 \cite{griffin2007caltech}. It takes the $256\times256\times3$ images as input which will be randomly cropped into $227\times227$ during the training and generates the 256-way softmax output for classes prediction. The ReLU activation function is not shown for brevity}
\label{table:caltech_cnn}
\begin{tabular}{|c|c|c|} \hline
		layer type 			&   kernel size/stride 		&    output size	\\
	\hline
		convolution 			& $11\times11/4$ 			&  $55\times55\times96$  \\
		max pool    			& $3\times3/2$ 				&  $27\times27\times96$  \\	
		LRN &              								&  $27\times27\times96$  \\
		convolution(padding2)		& $5\times5/1$ 				&  $27\times27\times256$  \\
		max pool 				& $3\times3/2$ 				&  $13\times13\times256$  \\
		LRN					& 						&  $13\times13\times256$  \\
		convolution(padding1)		& $3\times3/1$ 				&  $13\times13\times384$  \\
		convolution(padding1)		& $3\times3/1$ 				&  $13\times13\times384$  \\
		convolution(padding1)		& $3\times3/1$ 				&  $13\times13\times256$  \\
		max pool 				& $3\times3/2$ 				&  $6\times6\times256$  \\
		fc(dropout$50\%$) 		&  						&  $1\times1\times4096$  \\
		fc(dropout$50\%$) 		&  						&  $1\times1\times4096$  \\		
		softmax 				&             			 	&  $1\times1\times256$  \\
	\hline
\end{tabular}
\end{center}
\end{table}

%


\begin{figure*}
\begin{center}
\includegraphics[width=0.8 \textwidth]{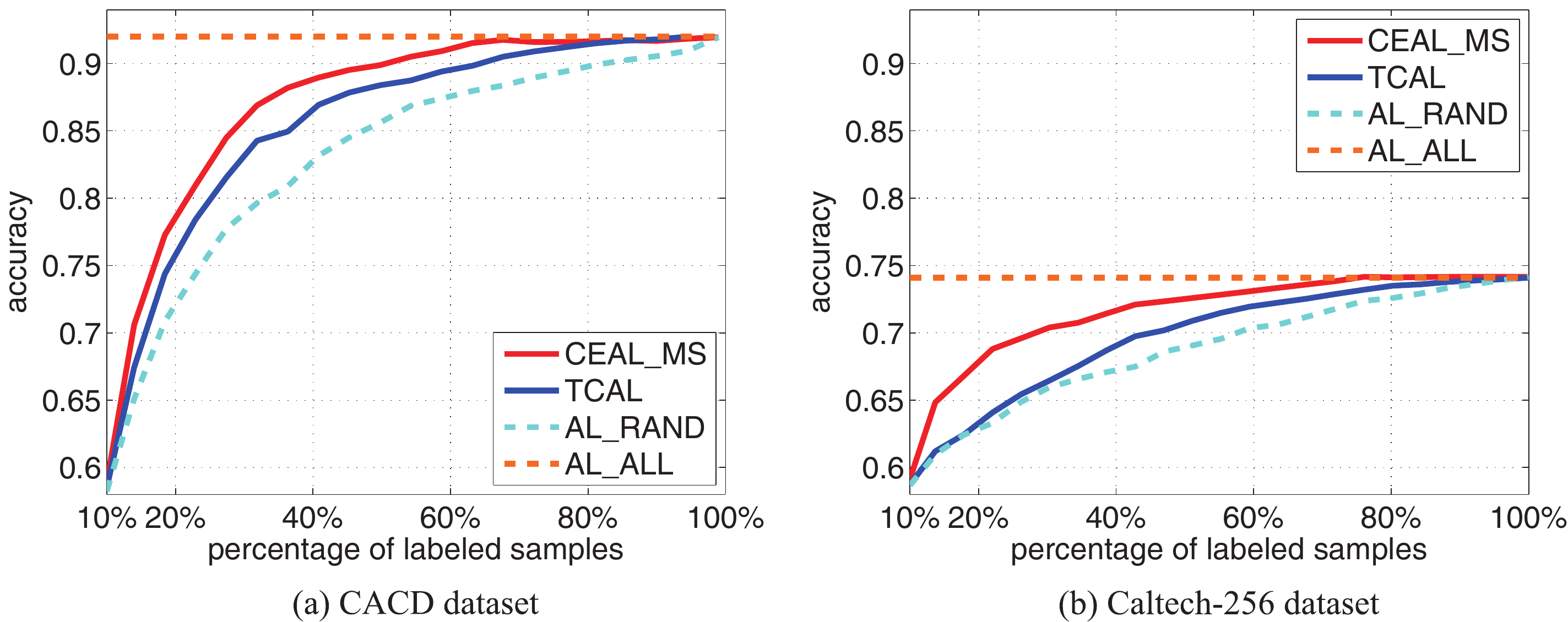}
\caption{Classification accuracy under different percentages of annotated samples of the whole training set on (a) CACD and (b) Caltech-256 datasets. Our proposed method CEAL\_MS performs consistently better than the compared TCAL and AL\_RAND.}
\label{fig:cealvstcal}
\end{center}
\end{figure*}

\begin{table*}
\caption{Class accuracy per some specific AL iterations on the CACD and Caltech-256 dataset.}
\centering
\subtable[CACD]{
\begin{tabular}{|c|c|c|c|c|c|c|c|c|c|c|c|c|c|c|} 
\hline
\hline
Training iteration & 0 & 2 & 4 & 6 & 8 & 10 & 12 & 14 & 16 & 18 & 20 \\
\hline
Percentage of labeled samples & 0.1 & 0.18 & 0.27 & 0.36 & 0.45 & 0.54 & 0.63 & 0.72 & 0.81 & 0.90 & 1 \\
\hline
CEAL\_MS & 57.4\% & 77.3\% & 84.5\% & 88.2\% & 89.5\% & 90.5\% & 91.5\% & 91.6\% & 91.7\% & 91.7\% & 92.0\% \\
TCAL & 57.4\% & 74.4\% & 81.6\% & 85.0\% & 87.9\% & 88.8\% & 89.8\% & 90.9\% & 91.5\% & 91.8\% & 91.9\% \\
AL\_RAND & 57.4\% & 70.9\% & 77.7\% & 80.9\% & 84.5\% & 86.9\% & 88.0\% & 89.0\% & 89.9\% & 90.6\% & 92.0\% \\
AL\_ALL & - & - & - & - & - & - & - & - & - & - & 92.0\% \\
\hline
\hline
\end{tabular}
\label{tab:cacd}
}
\qquad
\subtable[Caltech-256]{        
\begin{tabular}{|c|c|c|c|c|c|c|c|c|c|c|c|c|} 
\hline
\hline
Training iteration & 0 & 2 & 4 & 6 & 8 & 10 & 12 & 14 & 16 & 18 & 20 & 22 \\
\hline
Percentage of labeled samples & 0.10 & 0.18 & 0.26 & 0.34 & 0.43 & 0.51 & 0.59 & 0.68 & 0.76 & 0.84 & 0.93 & 1 \\
\hline
CEAL\_MS & 58.4\% & 64.8\% & 68.8\% & 70.4\% & 71.4\% & 72.3\% & 72.8\% & 73.3\% & 73.8\% & 74.2\% & 74.2\% & 74.2\% \\
TCAL & 58.4\% & 62.4\% & 65.4\% & 67.5\% & 69.8\% & 70.9\% & 71.9\% & 72.5\% & 73.2\% & 73.6\% & 73.9\% & 74.2\% \\
AL\_RAND & 58.4\% & 62.4\% & 64.8\% & 66.6\% & 67.5\% & 69.1\% & 70.4\% & 71.2\% & 72.4\% & 72.9\% & 73.6\% & 74.2\% \\
AL\_ALL & - & - & - & - & - & - & - & - & - & - & - & 74.2\% \\
\hline
\hline
\end{tabular}
\label{tab:caltech256}
}
\end{table*}

%

\par
We use different network architectures for CACD \cite{chen2014cross} and Caltech256 \cite{griffin2007caltech} datasets because the difference between face and object is relatively large. Table~\ref{table:cacd_cnn} shows the overall network architecture for CACD experiments and table~\ref{table:caltech_cnn} shows the overall network architecture for Caltech-256 experiments. We use the Alexnet \cite{krizhevsky2012imagenet} as the network architecture for Caltech-256 and using the ImageNet ILSVRC dataset \cite{DBLP:journals/corr/RussakovskyDSKSMHKKBBF14} pre-trained model as the starting point following the setting of \cite{DBLP:conf/icml/DonahueJVHZTD14}. Then we keep all layers fixed and just modify the last layer to be the 256-way softmax classifier to perform the Caltech-256 experiments. We employ Caffe  \cite{jia2014caffe} for CNN implementation. 

{
For CACD, we set the learning rates of all the layers as $0.01$. For Caltech-256, we set the learning rates of all the layers as $0.001$ except for the softmax layer which is set to $0.01$. All the experiments are conducted on a common desktop PC with an intel 3.8GHz CPU and a Titan X GPU. Average 17 hours are needed to finish training on the CACD dataset with 44708 images.

\subsubsection{Comparison Methods}
To demonstrate that our proposed CEAL framework can improve the classification performance with less labeled data, we compare CEAL with a new state-of-the-art active learning (TCAL) and baseline methods (AL\_ALL and AL\_RAND):

\begin{itemize}
\item AL\_ALL: We manually label all the training samples and use them to train the CNN. This method can be regarded as the upper bound (best performance that CNN can reach with all labeled training samples).
\item AL\_RAND: During the training process, we randomly select samples to be annotated to fine-tune the CNN. This method discards all active learning techniques and can be considered as the lower bound.
\item Triple Criteria Active Learning (TCAL)~\cite{DBLP:journals/tgrs/DemirB15}: TCAL is a comprehensive active learning approach and is well designed to jointly evaluate sample selection criteria (uncertainty, diversity and density), and has overcome the state-of-the-art methods  with much less annotations. TCAL represents those methods who intend to mine minority informative samples to improve the performance. Thus, we regard it as a relevant competitor. 
\end{itemize}

\begin{figure*}
\begin{center}
\includegraphics[width=1 \textwidth]{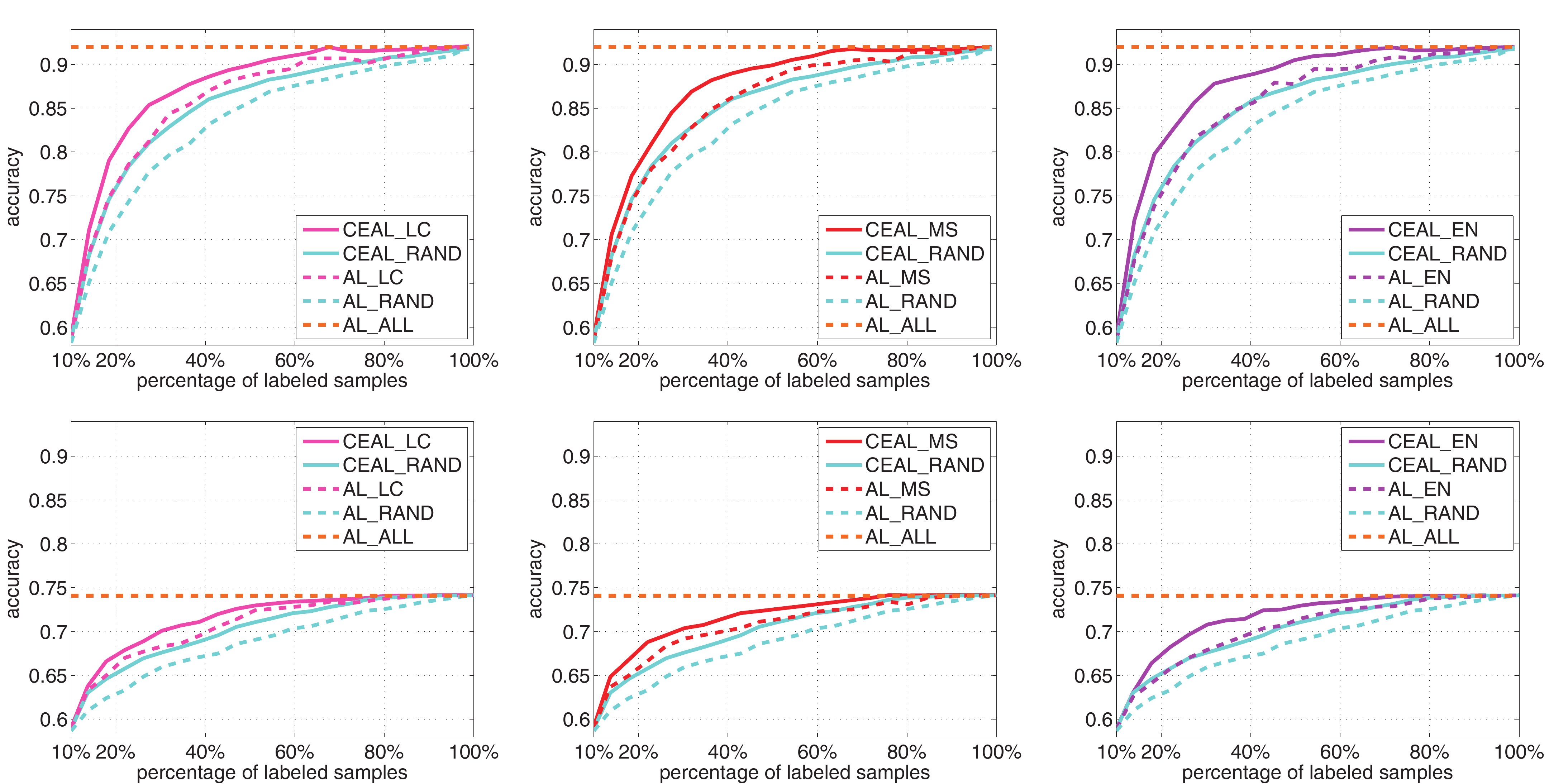}
\caption{Extensive study for different informative sample selection criteria on CACD (the first row) and Caltech-256 (the second row) datasets. These criteria includes least confidence (LC, the first column), margin sampling (MS, the second column) and entropy (EN, the third column). One can observe that our CEAL framework works consistently well on the common information sample selection criteria.}
\label{fig:analy_component}
\end{center}
\end{figure*}

\begin{figure*}
\begin{center}
\includegraphics[width=0.8 \textwidth]{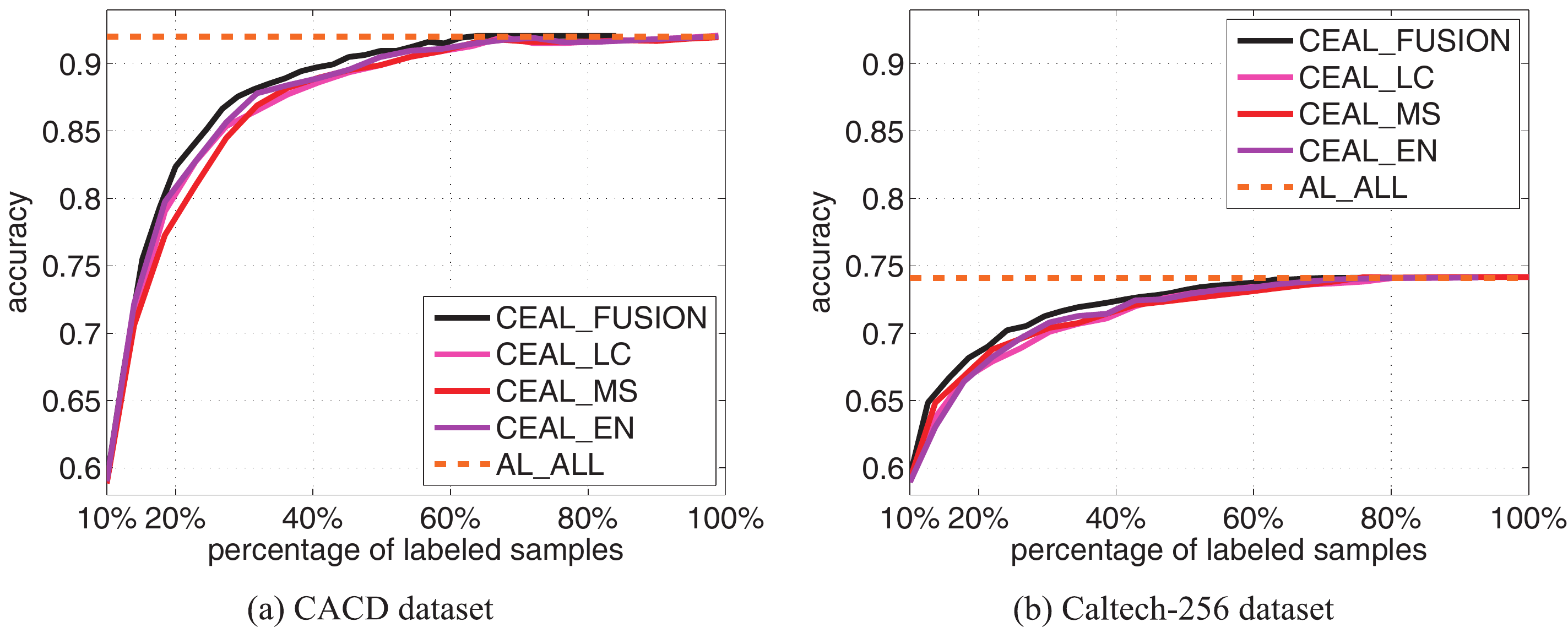}
\caption{Comparison between different informative sample selection criteria and their fusion (CEAL\_FUSION) on CACD (left) and Caltech-256 (right) datasets.}
\label{fig:cmp_component}
\end{center}
\end{figure*}

{\em Implementation Details.} The compared methods share the same CNN architecture with our CEAL on the both datasets. The only difference in the sample selection criteria. 
For the BaseLine method, we select all training samples to fine-tune the CNN, i.e. all labels are used. For TCAL, we follow the pipeline of~\cite{DBLP:journals/tgrs/DemirB15} by training a SVM classifier and then applying the uncertainty, diversity and density criteria to select the most informative samples. Specifically, the uncertainty of samples is assessed according to the margin sampling strategy. The diversity is calculated by clustering the most uncertain samples via k-means with histogram intersection kernel. The density of one sample is measured by calculating the average distance with other samples within a cluster it belonged to. For each cluster, the highest density (i.e., the smallest average distance) sample is selected as the most informative sample. 
For CACD, we cluster 2000 most uncertain samples and select 500 most informative samples according to above mentioned diversity and density. For Caltech-256, we select 250 most informative samples from 1000 most uncertain samples. To make a fair comparison, samples selected in each iteration by the TCAL are also used to fine-tune the CNN to learn the optimal feature representation as AL\_RAND. Once optimal feature learned, the SVM classifier of TCAL is further updated.

\subsection{Comparison Results and Empirical Analysis}

\subsubsection{Comparison Results}
To demonstrate the effectiveness of our proposed framework, we also apply margin sampling criterion to measure the uncertainty of samples and denote this method as CEAL\_MS. 
Fig.~\ref{fig:cealvstcal} illustrates the accuracy-percentage of annotated samples curve of AL\_RAND, AL\_ALL, TCAL and the proposed CEAL\_MS on both CACD and Caltech-256 datasets. This curve demonstrates the classification accuracy under different percentages of annotated samples of the whole training set.

As illustrated in Fig.~\ref{fig:cealvstcal}, Tab.~\ref{tab:cacd} and Tab.~\ref{tab:caltech256}, our proposed CEAL framework overcomes the compared method from the aspects of the recognition accuracy and user annotation amount. 
From the aspect of recognition accuracy, given the same percentage of annotated samples, our CEAL\_MS outperforms the compared method in a clear margin, especially when the percentage of annotated samples is low.
From the aspect of the user annotation amount, to achieve 91.5\% recognition accuracy on the CACD dataset, AL\_RAND and TCAL require 99\% and 81\% labeled training samples, respectively. CEAL\_MS needs only 63\% labeled samples and reduces around 36\% and 18\% user annotations, compared to AL\_RAND and TCAL. To achieve the 73.8\% accuracy on the caltech-256 dataset, AL\_RAND and TCAL require 97\% and 93\% labeled samples, respectively. CEAL\_MS needs only 78\% labeled samples and reduces around 19\% and 15\% user annotations, compared to AL\_RAND and TCAL. This justifies that our proposed CEAL framework can effectively reduce the need of labeled samples. 

From above results, one can see that our proposed frame CEAL performs consistently better than the state-of-the-art method TCAL in both recognition accuracy and user annotation amount through fair comparisons. This is due to that TCAL only mines minority informative samples and is not able to provide sufficient training data for feature learning under deep image classification scenario. Hence, our CEAL has a competitive advantage in deep image classification task.  
To clearly analyze our CEAL and justify the effectiveness of its component, we have conducted the several experiments and discussed in the following subsections.

\subsubsection{Component Analysis}
To justify that the proposed CEAL can works consistently well on the common informative sample selection criteria, we implement {three variants of CEAL according to least confidence (LC), margin sampling (MS) and entropy (EN) to assess uncertain samples. These three variants are denoted as CEAL\_LC, CEAL\_MS and CEAL\_EN.} Meanwhile, to show the raw performance of these criteria, we discard the cost-effective high confidence sample selection of above mentioned variants and denoted the discarded versions as AL\_LC, AL\_MS and AL\_EN. To clarify the contribution of our proposed pseudo-labeling majority high confidence sample strategy, we further introduce this strategy into the AL\_RAND and denote this variant as CEAL\_RAND. Since AL\_RAND means randomly select samples to be annotated, CEAL\_RAND reflects the original contribution of the pseudo-labeled majority high confidence sample strategy, i.e., CEAL\_RAND denotes the method who only uses pseudo-labeled majority samples.

Fig.~\ref{fig:analy_component} illustrates the results of these variants on dataset CACD (the first row) and Caltech-256 (the second row). The results demonstrate that giving the same percentage of labeled samples and compared with AL\_RAND, CEAL\_RAND, simply exploiting pseudo-labeled majority samples, obtains similar performance gain as AL\_LC, AL\_MS and AL\_EN, which employs informative sample selection criterion. This justifies that our proposed pseudo-labeling majority sample strategy is effective as some common informative sample selection criteria. Moreover, as one can see that in Fig.~\ref{fig:analy_component}, CEAL\_LC, CEAL\_MS and CEAL\_EN all consistently outperform the pure pseudo-labeling samples version CEAL\_RAND and their excluding pseudo-labeled samples versions AL\_LC, AL\_MS and AL\_EN in a clear margin on both the CACD and Caltech-256 datasets, respectively. This validates that our proposed pseudo-labeling majority sample strategy is complementary to the common informative sample selection criteria and can further significantly improve the recognition performance.

\begin{figure}[ht]
\begin{center}
\includegraphics[width=0.4 \textwidth]{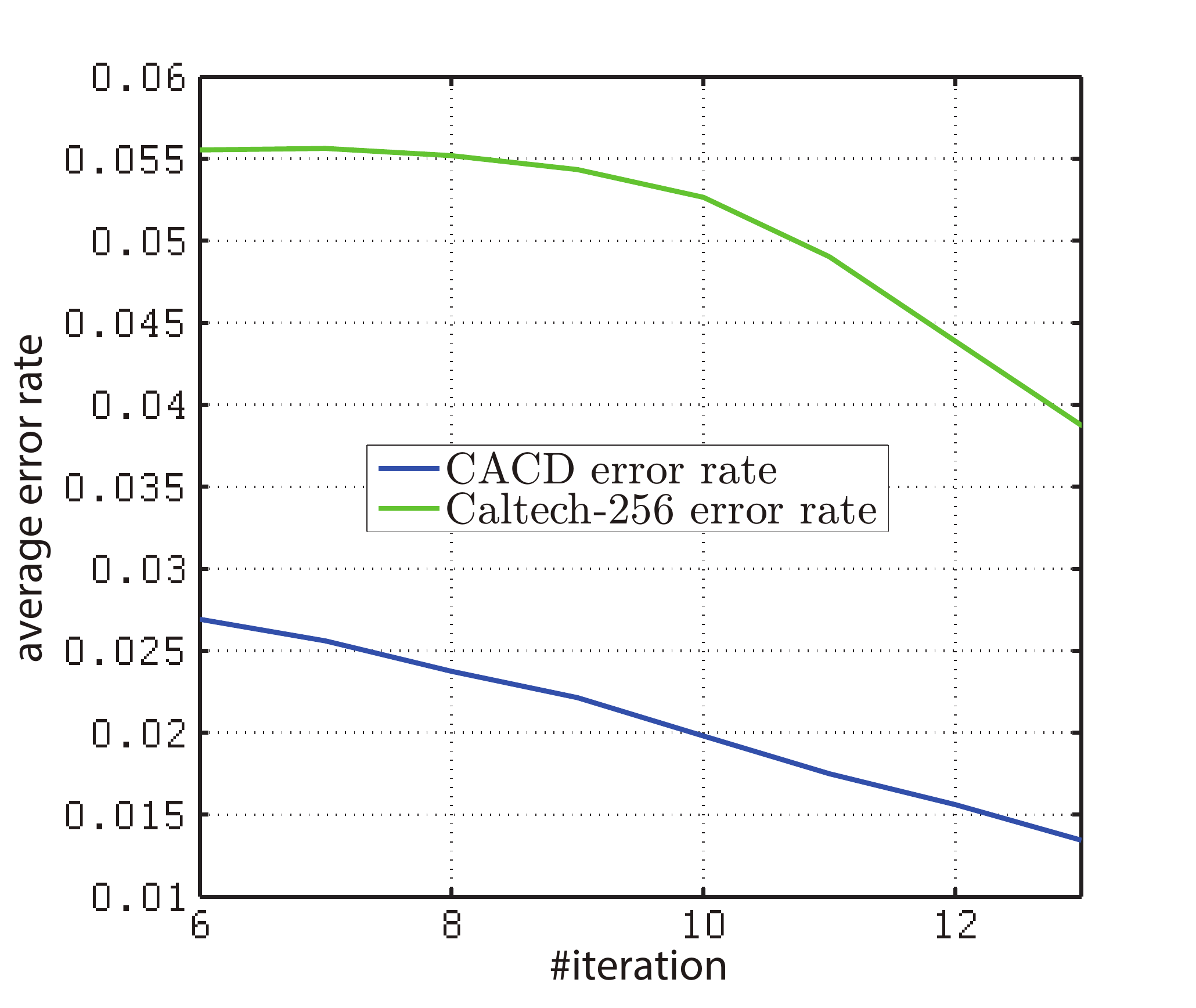}
\vspace{-4mm}
\caption{The average error rate of the pseudo-labels of high confidence samples assigned by the heuristic strategy on CACD and Caltech-256 datasets experiments. The vertical axes represent the average error rate and the horizontal axes represent the learning iteration. Our proposed CEAL framework can assign reliable pseudo-labels to the unlabeled samples under acceptable average error rate}
\label{fig:error}
\end{center}
\end{figure}

To analyze the choice of informative sample selection criteria, we have made a comparison among three above mentioned criteria. We also make an attempt to simply combine them together. Specifically, in each iteration, we select top $K/2$ samples according to each criterion respectively. Then we remove repeated ones (i.e., some samples may be selected by the three criteria at the same time) from obtained $3K/2$ samples. After removing the repeated samples, we randomly select $K$ samples from them to require user annotations. We denote this method as CEAL\_FUSION.

Fig.~\ref{fig:cmp_component} illustrates that CEAL\_LC, CEAL\_MS and CEAL\_EN have similar performance while CEAL\_FUSION performs better. This demonstrates that informative sample selection criterion still plays an important role in improving the recognition accuracy. Though being a minority, the informative samples have great potential impact on the classifier.

\subsection{Reliability of CEAL}
From the above experiments, we know that the performance of our framework is better from other methods, which shows the superiority of introducing the majority pseudo-labeled samples. But how does the accuracy of assigning the pseudo-label to those high confidence samples? In order to demonstrate the reliability of our proposed CEAL framework, we also evaluate the average error in selecting high confidence samples. Fig.~\ref{fig:error} shows the error rate of assigning pseudo-label along with the learning iteration. As one can see that, the average error rate is quite low (say less than 3\% on the CACD dataset and less than 5.5\% on the Caltech-256 dataset) even at early iterations. Hence, our proposed CEAL framework can assign reliable pseudo-labels to the unlabeled samples under acceptable average error rate along with the learning iteration.

\begin{figure}[ht]
\begin{center}
\includegraphics[width=0.8 \columnwidth]{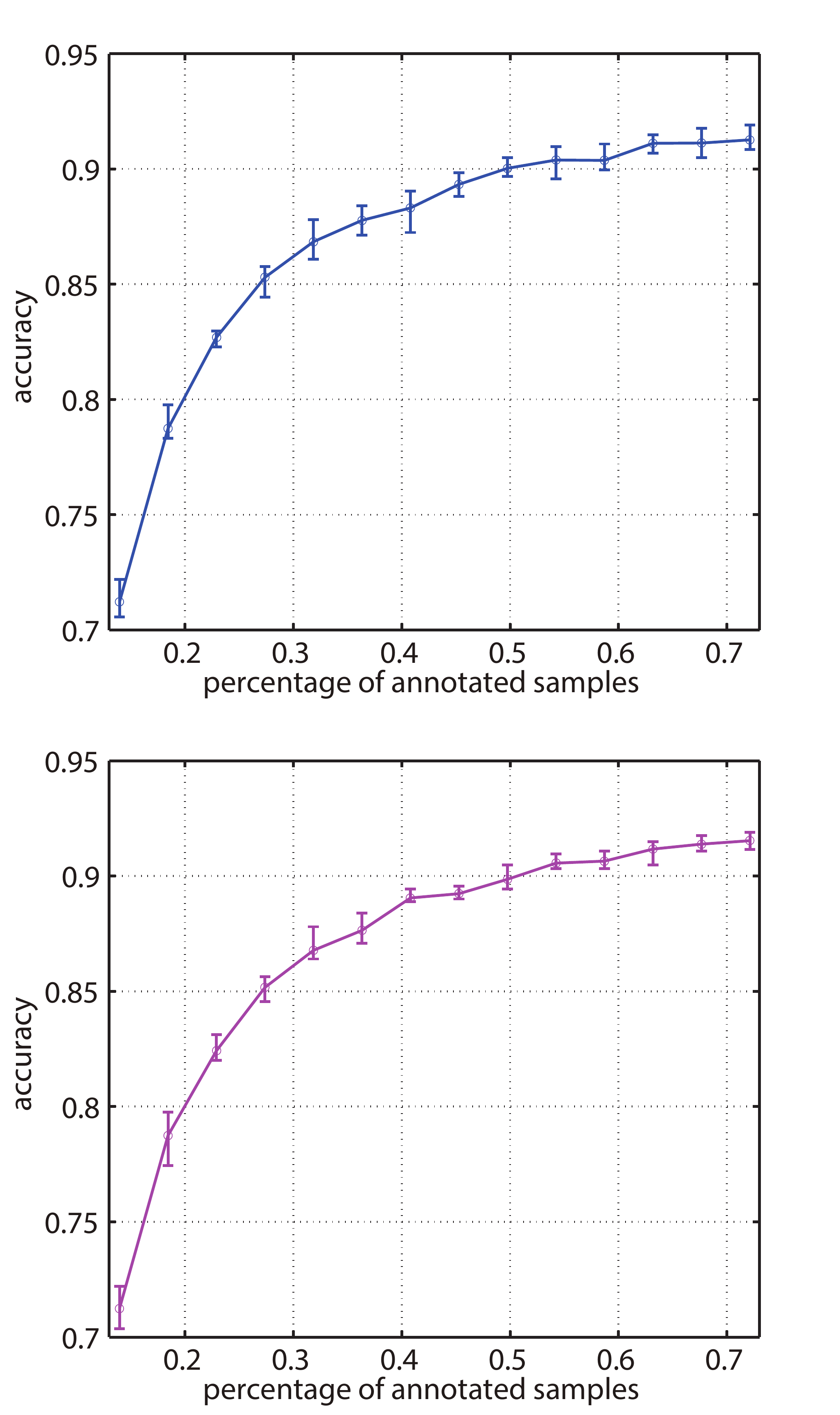}
\vspace{-4mm}
\caption{{ The sensitivity analysis of heuristic threshold $\delta$ (the first row) and decay rate $dr$ (the second row). One can observe that these parameters do not substantially affect the overall system performance.}}
\label{fig:decay_sensity}
\end{center}
\end{figure}

\subsection{Sensitivity of High Confidence Threshold}
Since the training phase of deep convolutional neural networks is time-consuming, it is not affordable to employ a try and error approach to set the threshold for defining high confidence samples. We further analyze the sensitivity of the threshold parameters $\delta$ (threshold) and $dr$ (threshold decay rate) on our system performance on CACD dataset using the CEAL\_EN. While analyzing the sensitivity of parameter $\delta$ on our system, we fix the decrease rate $dr$ to 0.0033. We fix the threshold $\delta$ to 0.05 when analyzing the sensitivity of $dr$. Results of the sensitivity analysis of $\delta$ (range 0.045 to 0.1) is shown in the first row of Fig.~\ref{fig:decay_sensity}, while the sensitivity analysis of $dr$ (range 0.001 to 0.0035) is shown in the second row of Fig.~\ref{fig:decay_sensity}. Note that the test range of $\delta$ and $dr$ is set to ensure the majority high confidence assumption of this paper. Though the range of \{$\delta$, $dr$\} seems to be narrow from the value, it leads to the significant difference: about 10\%~60\% samples are pseudo-labeled in high confidence sample selection.
Lower standard deviation of the accuracy in Fig.~\ref{fig:decay_sensity} proves that the choice of these parameters does not significantly affect the overall system performance.
}

\section{Conclusions}
In this paper, we propose a cost-effective active learning framework for deep image classification tasks, which employs a complementary sample selection strategy: Progressively select minority most informative samples and pseudo-label majority high confidence samples for model updating. In such a holistic manner, the minority labeled samples benefit the decision boundary of classifier and the majority pseudo-labeled samples provide sufficient training data for robust feature learning. Extensive experiment results on two public challenging benchmarks justify the effectiveness of our proposed CEAL framework. In future works, we plan to apply our framework on more challenging large-scale object recognition tasks (e.g., 1000 categories in ImageNet). And we plan to incorporate more persons from the CACD dataset to evaluate our framework. Moreover, we plan to generalize our framework into other multi-label object recognition tasks (e.g., 20 categories in PASCAL VOC).

\bibliographystyle{IEEEtran}

\begin{IEEEbiography}[{\includegraphics[width=1in,height=1.25in,clip,keepaspectratio]{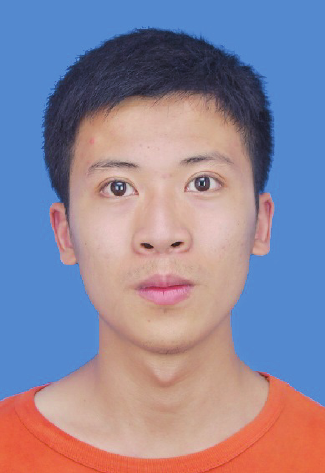}}]{Keze Wang}
received the BS degree in software engineering from Sun Yat-Sen University, Guangzhou, China, in 2012. He is currently pursuing a double Ph.D. degree in computer science and technology at Sun Yat-Sen University and Hong Kong Polytechnic University, advised by Professor Liang Lin and Lei Zhang. His current research interests include computer vision and machine learning.
\end{IEEEbiography}

\begin{IEEEbiography}[{\includegraphics[width=1in,height=1.25in,clip,keepaspectratio]{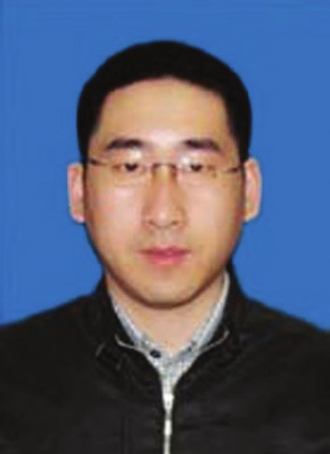}}]{Dongyu Zhang}  is a research associate with school of data and computer science, Sun-Yat-Sen University (SYSU). He received his B.S. degree and Ph.D. in Harbin Institute of Technology University, Heilongjiang, China, in 2003 and 2010. His current research interests include computer vision and machine learning. 
\end{IEEEbiography}

\begin{IEEEbiography}[{\includegraphics[width=1in,height=1.25in,clip,keepaspectratio]{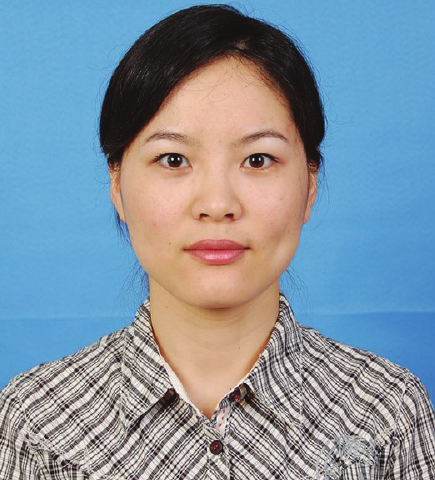}}]{Ya Li} is a lecturer in School of Computer Science and Educational Software at Guangzhou University, China. She received the B.E. degree from Zhengzhou University, Zhengzhou, China, in 2002, M.E. degree from Southwest Jiaotong University, Chengdu, China, in 2006 and Ph.D. degree from Sun Yat-sen University, Guangzhou, China, in 2015. Her research focuses on computer vision and machine learning.
\end{IEEEbiography}

\begin{IEEEbiography}[{\includegraphics[width=1in,height=1.25in,clip,keepaspectratio]{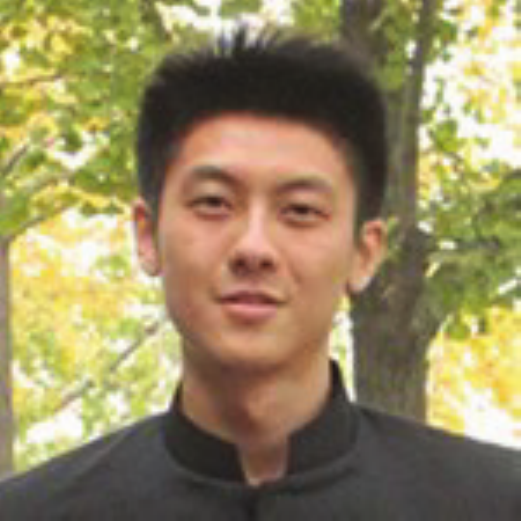}}]{Ruimao Zhang} received the B.E. degree in the School of Software from Sun Yat-Sen University (SYSU) in 2011. Now he is a Ph.D. candidate of Computer Science in the School of Information Science and Technology, Sun Yat-Sen University, Guangzhou, China. From 2013 to 2014, he was a visiting Ph.D. student with the Department of Computing, Hong Kong Polytechnic University (PolyU). His current research interests are computer vision, pattern recognition, machine learning and related applications.
\end{IEEEbiography}


\begin{IEEEbiography}[{\includegraphics[width=1in,height=1.25in,clip,keepaspectratio]{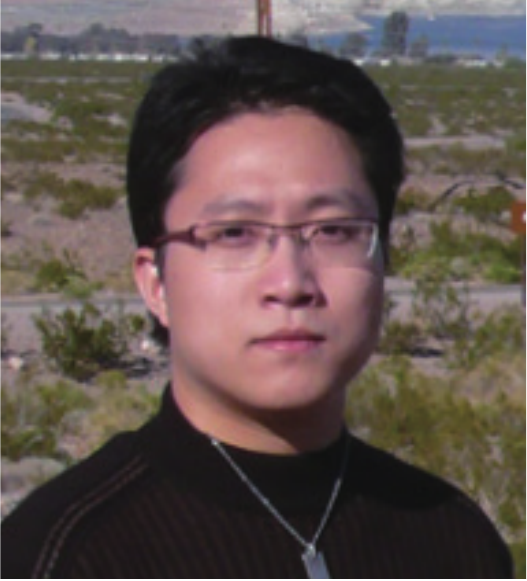}}]{Liang Lin} is a Professor with the School of computer science, Sun Yat-Sen University (SYSU), China. He received the B.S. and Ph.D. degrees from the Beijing Institute of Technology (BIT), Beijing, China, in 1999 and 2008, respectively. From 2008 to 2010, he was a Post-Doctoral Research Fellow with the Department of Statistics, University of California, Los Angeles UCLA. He worked as a Visiting Scholar with the Department of Computing, Hong Kong Polytechnic Uni versity, Hong Kong and with the Department of Electronic Engineering at the Chinese University of Hong Kong. His research focuses on new models, algorithms and systems for intelligent processing and understanding of visual data such as images and videos. He has published more than 100 papers in top tier academic journals and conferences. He currently serves as an associate editor of IEEE Tran. Human-Machine Systems. He received the Best Paper Runners-Up Award in ACM NPAR 2010, Google Faculty Award in 2012, Best Student Paper Award in IEEE ICME 2014, and Hong Kong Scholars Award in 2014.
\end{IEEEbiography}

\end{document}